# Deep Learning-based Anomaly Detection and Log Analysis for Computer Networks

**Shuzhan Wang[1], Ruxue Jiang[2], Zhaoqi Wang[3], Yan Zhou[4*]**
[1]School of Information Sciences, University of Illinois Urbana-Champaign, Champaign, Illinois, 61820, United States; zzsusan.wang@gmail.com
[2]Northeastern University, Seattle, Washington, 98109; snow0911@outlook.com
[3]Viterbi School of Engineering, University of Southern California, Los Angeles, California, 90089, United States;josiewangwzq@gmail.com
[4]Northeastern University San Jose, California, 95131, United States; yanzhouca23@gmail.com
*Corresponding Author: yanzhouca23@gmail.com

## ABSTRACT

Computer network anomaly detection and log analysis, as an important topic in the field of network security, has been a key task to ensure network security and system reliability. First, existing network anomaly detection and log analysis methods are often challenged by high-dimensional data and complex network topologies, resulting in unstable performance and high false-positive rates. In addition, traditional methods are usually difficult to handle time-series data, which is crucial for anomaly detection and log analysis. Therefore, we need a more efficient and accurate method to cope with these problems. To compensate for the shortcomings of current methods, we propose an innovative fusion model that integrates Isolation Forest, GAN (Generative Adversarial Network), and Transformer with each other, and each of them plays a unique role. Isolation Forest is used to quickly identify anomalous data points, and GAN is used to generate synthetic data with the real data distribution characteristics to augment the training dataset, while the Transformer is used for modeling and context extraction on time series data. The synergy of these three components makes our model more accurate and robust in anomaly detection and log analysis tasks. We validate the effectiveness of this fusion model in an extensive experimental evaluation. Experimental results show that our model significantly improves the accuracy of anomaly detection while reducing the false alarm rate, which helps to detect potential network problems in advance. The model also performs well in the log analysis task and is able to quickly identify anomalous behaviors, which helps to improve the stability of the system. The significance of this study is that it introduces advanced deep learning techniques, which brings new ideas to the field of computer network anomaly detection and log analysis. In addition, the successful application of the Isolation Forest-GAN-Transformer model also helps to promote the development of deep learning in the field of network security, which provides valuable references and lessons for future research. In the future work, the model will be further optimized and more deep learning methods will be explored to meet the evolving network threat and log analysis needs to ensure the security and availability of networks and systems.

Keywords: Anomaly Detection, Log Analysis, Network, Isolation Forest, GAN, Transformer

## 1. Introduction

Computer network anomaly detection and log analysis has attracted widespread attention as a key issue in the field of information technology[1]. In today's digitalized society, the rapid development and wide application of computer networks have made them an indispensable part of business, government and personal life. However, with the increasing size and complexity of networks, anomalous behaviors and security threats in network systems have become increasingly severe and complex, and the frequency and impact of such network anomalous events have become increasingly significant[2]. Network anomalous events can lead to serious consequences such as service disruption, data leakage, malicious attacks, and system crashes, making anomaly detection critical in computer network security and management.

Anomaly detection, also known as anomaly detection or abnormal behavior detection, is a technology aimed at identifying events or data that deviate from expected behavior[3]. In the field of computer networks, anomalies may include malicious intrusions, network failures, performance degradation, or other unusual network behaviors[4]. These abnormal events can have a significant impact on the security, availability, and performance of networks. In recent years, the rapid development of deep learning technology has provided new opportunities to address anomaly detection challenges[5]. Deep learning is a machine learning method based on artificial neural networks, which can extract abstract features from large-scale and complex data, thereby improving the identification of abnormal behavior[6]. In the field of network security and performance management, researchers have started to adopt deep learning methods to enhance the effectiveness of anomaly detection and log analysis. These methods include Convolutional Neural Networks (CNNs), Recurrent Neural Networks (RNNs), and more advanced variants, which have achieved significant results in identifying abnormal behavior[7]. It is worth noting that in computer network anomaly detection and log analysis, time series data plays a crucial role because network data is typically recorded in chronological order. Therefore, time series forecasting becomes a key element of the research. By modeling and analyzing time series data using deep learning methods, we can more accurately predict future network behavior, promptly identify anomalies[8], and take necessary measures to maintain network security and performance stability.

However, although deep learning techniques with time series analysis hold new promise for anomaly detection, they still face many challenges in practice[9]. These challenges include dealing with high dimensional data, noise interference, data imbalance, and conceptual drift[10]. Traditional rule-based or statistical methods often struggle to address these issues effectively, and more advanced techniques are needed to overcome these challenges[11]. Based on the shortcomings of the previous work, we propose the Isolation Forest-GAN-Transformer network, which aims to further improve the accuracy and efficiency of anomaly detection and log analysis in computer networks. Our model combines several key elements such as Isolation Forest, Generative Adversarial Network (GAN), and Transformer to cope with some of the limitations of existing approaches.

First, Isolation Forest, as a tree structure-based anomaly detection method, is used to detect anomalous behaviors in networks quickly and effectively. Its unique isolation strategy can effectively separate anomalous data points, but it still has challenges in dealing with high-dimensional time series data and complex network structures. To further improve the performance of the model, we introduce GAN. GAN improves the robustness of the model by generating artifacts of real data, which is crucial for the performance of deep learning models in anomaly detection tasks. Additionally, we introduce the Transformer architecture for modeling time-series data. Transformer has been highly successful in areas such as natural language processing, and its self-attention mechanism allows it to capture long-term dependencies in time series, which is useful in network anomaly detection. By introducing Transformer into our modeling, we expect to be able to more accurately capture anomalous patterns and trends in network data.

The significance and advantage of our model is that it synthesizes several advanced techniques to overcome the limitations of traditional methods and improve the accuracy and efficiency of anomaly detection. With Isolation Forest, we are able to quickly identify anomalous data points. With GAN, we improve the robustness of the model so that it can handle complex network data. With Transformer, we are able to better capture patterns and correlations in time-series data, which improves the predictive performance of the model. The application potential of this comprehensive approach lies in improving network security, performance management, and log analysis, providing strong support for building more secure and reliable computer networks. By delving into the application of deep learning in computer network anomaly detection and log analysis, we will be able to better understand and solve the challenges we currently face and provide more reliable solutions for network security and performance management. In the subsequent part of this thesis, we will present our model architecture and experimental results in detail to verify its effectiveness in the field of anomaly detection and log analysis.

The contribution points of this paper are as follows:

1) Improve network security: Our research provides a more accurate and efficient solution for computer network anomaly detection and log analysis. This will help detect potential network threats, such as malicious intrusions or anomalous behaviors, in a timely manner, thus improving network security. This contribution has practical implications for the protection of critical infrastructure, sensitive data, and personal privacy.

2) Optimize performance management: With deep learning technology, our research enables us to analyze network performance data more accurately, identify performance problems and provide targeted solutions. This helps to improve network availability and efficiency, reduce potential service interruptions, lower O&M costs, and is of practical help to the normal operation of businesses and organizations.

3) Simplify Log Analysis: Our model automates the log analysis process to quickly detect and handle abnormal logs. This will save administrators' time and effort, allowing them to focus more on network management and security tasks, and helping to improve IT O&M efficiency.

## 2. Related Work

### 2.1 Anomaly Detection Methods Based on Traditional Machine Learning

In the field of network anomaly detection, traditional machine learning methods have been dominant[12]. One common approach is to build anomaly detection models using Support Vector Machines (SVMs)[13], which are able to separate normal and anomalous data points by finding the best hyperplane for the data[14]. Another common approach is statistically based methods such as Isolation Forest, which isolates anomalous data by building trees[15]. While these methods perform well in some scenarios, they may not work well when dealing with complex, high-dimensional network data.

However, traditional machine learning methods usually rely on manual feature engineering, which requires specialized knowledge to select and extract features for complex network data[16]. In addition, such approaches may not be flexible enough to cope with changing cyber threats and anomaly patterns.

### 2.2 Network Anomaly Detection Based on Time Series Analysis

Time series analysis has been an important research direction in the field of network anomaly detection and log analysis[17]. Researchers use the temporal nature of time series data to identify anomalous behavior. One common approach is to use autoregressive integrated moving average models (ARIMA) or seasonal decomposition methods to analyze the trend[18], seasonality, and residual components of time series data. These methods can be used to detect network performance degradation, anomalous events, and failures[19]. Another time-series based approach utilizes frequency domain analysis techniques such as Fourier transform and wavelet transform. These methods can be used to detect periodic anomalous behavior or unusual patterns in the frequency domain[20].

The shortcoming of this approach is that time series methods typically rely on assumptions about the stability and smoothness of the data, whereas real-world network data often contains noise and non-smoothness[21]. In addition, these methods may have difficulty with high-dimensional and large-scale time series data because they require the computation of complex mathematical operations.

### 2.3 Anomaly Detection Based on Log Analysis

Log analysis is an important research area in the field of network anomaly detection and log analysis[22]. Log files record the activities and events of systems such as network devices, servers, applications, etc., and thus provide important information about the state of the system[20]. Researchers use the information in log files to detect anomalous behavior. A common approach is to use rule-based log analysis. This approach relies on predefined sets of rules for detecting events that do not match those rules. For example, if unauthorized access attempts or unusual system error messages appear in the logs, a system administrator can trigger an alert using a rule[23]. Another approach is to use natural language processing (NLP) techniques to analyze log text[24]. Researchers have developed NLP-based models that can recognize key information in log text, such as descriptions of anomalous events or error messages. This approach can help automate the anomaly detection process.

However, although this approach is widely used, it still has some drawbacks. For example, rule-based methods usually require manual definition of rules, which may lead to missed or false alarms. And NLP-based methods may be affected by the quality and diversity of log text, may require a large amount of labeled data for training, and have limited ability to generalize to different types of log data[25].

### 2.4 Automated Network Management and Fault Detection

Automated network management is a critical task aimed at ensuring the reliability and performance of networks. In this area, researchers and engineers have been exploring various methods and techniques to improve the management and fault detection of networks[26]. One common approach is rule-based network management. This approach relies on predefined sets of rules for detecting anomalous behavior or potential problems in a network[27]. For example, rules can be used to detect unusual patterns in network traffic, faulty devices, or potential network attacks. Another approach is to use machine learning techniques, such as decision trees, random forests, or support vector machines, to build network anomaly detection models[28]. These models can analyze large amounts of network data, identify anomalous behavior, and provide real-time alerts. In addition, deep learning methods, such as Convolutional Neural Networks (CNN) and Recurrent Neural Networks (RNN), have been applied in the field of network management to improve the accuracy of anomaly detection[29].

However, rule-based methods may require a lot of manual definition and maintenance, while machine learning and deep learning methods usually require a large amount of labeled data for training. In addition, the performance and generalization ability of the model may be affected by data quality and noise. As a result, researchers are constantly looking for ways to improve automated network management in order to increase network availability and performance.

## 3. Materials and Methods

### 3.1 Overview of Our Network

To cope with evolving computer network threats, we propose an innovative deep learning approach that fuses key components such as Isolation Forest (IF), Generative Adversarial Network (GAN), and Transformer into a new integrated model. The design concept of this model aims to enable computer network anomaly detection and log analysis to achieve superior performance and accuracy in increasingly complex threat environments by integrating diverse technologies. Its core idea lies in organically combining these components to more effectively capture and respond to a variety of network security challenges, ensuring the reliability and security of computer networks. This innovative deep learning approach represents an important contribution to the field of cybersecurity, providing a powerful tool and methodology to address future cyber threats.

First, the raw computer network dataset is fed into the Isolation Forest component. The task of the Isolation Forest is to quickly separate normal network traffic data from potentially anomalous data points by constructing a randomized binary tree. The purpose of this step is to generate an anomaly score for each data point, where higher scores indicate potentially anomalous data points. Also, the GAN works in conjunction with the Isolation Forest component, which consists of two parts, the generator and the discriminator. The task of the generator is to generate synthetic data similar to normal network traffic data to approximate the distribution of real data. The discriminator, on the other hand, evaluates the similarity between the generated data and the real data and pushes the generator to generate more realistic data. This process is iterative, and the generator continuously improves the quality of the generated data, enhancing the fidelity of the data while helping to capture potentially anomalous data features. The generated synthetic data and raw web log data are fed into the Transformer model in a time-series fashion. The Transformer's task is to learn the temporal features in the data, including dependencies and trends between events. This is achieved through efficient time-series modeling, which allows Transformer to analyze weblog data more comprehensively, both the generated synthetic data and the raw data, thus improving the accuracy of anomaly detection. Ultimately, the results of Isolation Forest's initial screening, GAN's data enhancement, and Transformer's temporal modeling are combined for comprehensive anomaly detection and log analysis. The generated data and raw data can be compared, and anomaly scores can be compared to the model's thresholds to identify potential anomalous behavior. The synergy of the entire model makes it possible to capture and identify anomalous behaviors in computer networks more accurately, improving network security.

The overall model structure is shown in Fig 1. This graphical representation reflects the complex yet efficient working mechanism of the Isolation Forest-GAN-Transformer model. The model is based on the complementary collaboration of several key components, providing a comprehensive and highly adaptive solution to the problem of anomaly detection and log analysis in computer networks. Overall, the integrated model we study not only integrates the strengths of different technologies, but also fully exploits the synergies among them to improve the accuracy and efficiency of computer network anomaly detection and log analysis. The potential of this model is not only limited to processing large-scale network data, but also brings important innovations to the field of cybersecurity and provides strong support for dealing with evolving cyber threats. Our research represents an important contribution to the field of cybersecurity, providing key tools and methods to ensure the security and stability of computer networks. This is a promising and strategically important research endeavor for a wide range of applications.

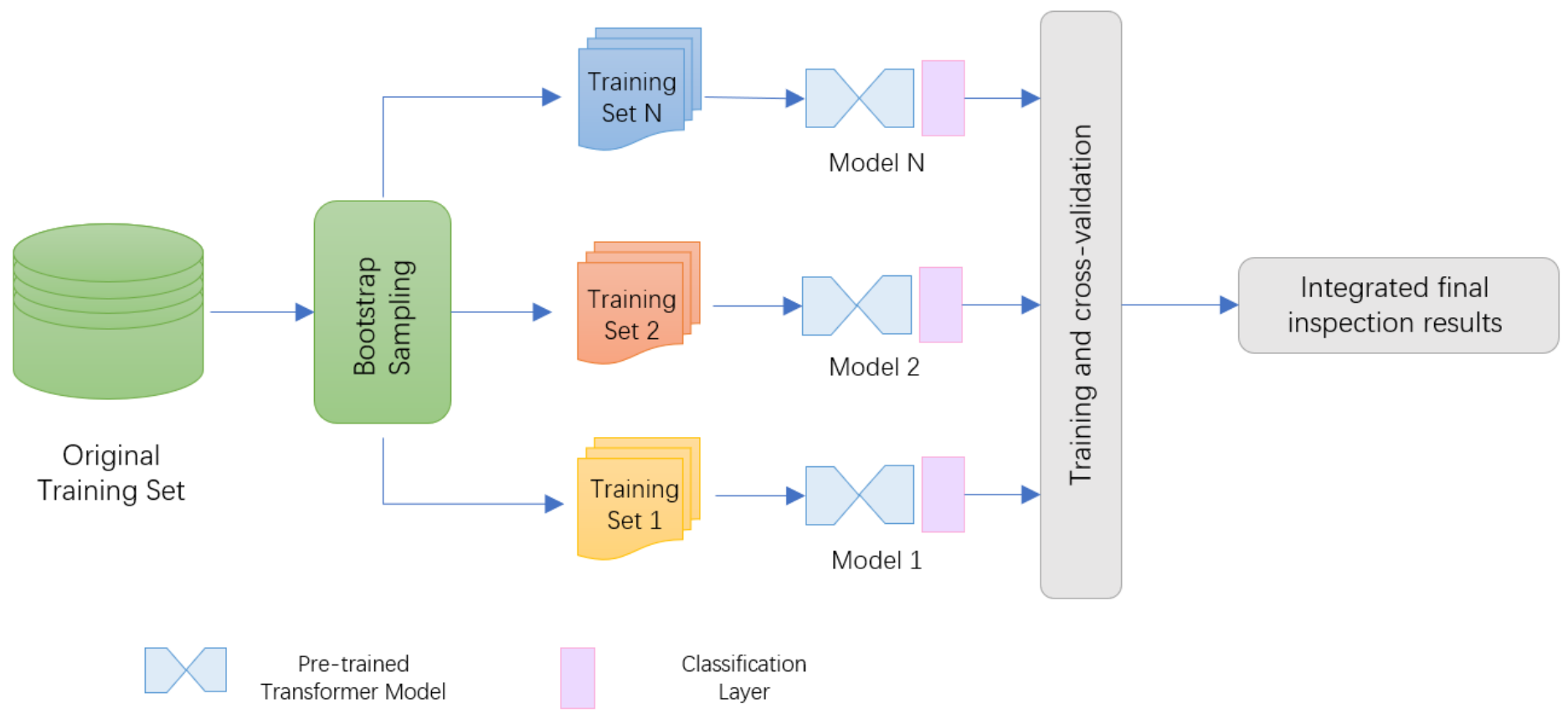

Figure 1. The structure of our model.

### 3.2 Isolation Forest

The Isolation Forest algorithm, conceptualized by Feier Liu and colleagues in 2008, represents a significant advancement in anomaly detection, distinguishing itself through its remarkable efficiency and intuitive approach. At the heart of this algorithm lies the principle that anomalies are inherently more "isolated" within a dataset, primarily due to their scarcity and distinctive properties relative to normal data points. The algorithm embarks on its task by constructing a 'forest' composed of numerous trees, each generated via random partitioning of the data. This entails the arbitrary selection of a feature and a corresponding threshold value, which together dictate the division of the dataset. Such divisions persist recursively, ultimately isolating each data point either into a unique node or until a tree reaches its maximum predetermined depth.

The essence of the Isolation Forest's effectiveness is encapsulated in its method for calculating an anomaly score for each data point, derived from the average length of the path leading to its isolation. Contrary to normal data points, which typically necessitate numerous splits—thereby incurring longer path lengths, anomalous data points are characterized by notably shorter paths, facilitating their swift identification. This scoring mechanism is a testament to the algorithm's unparalleled efficiency, particularly evident when processing datasets of considerable scale and complexity. The structure of the Isolation Forest is visually summarized in Figure 2, providing a clear depiction of its operational framework.

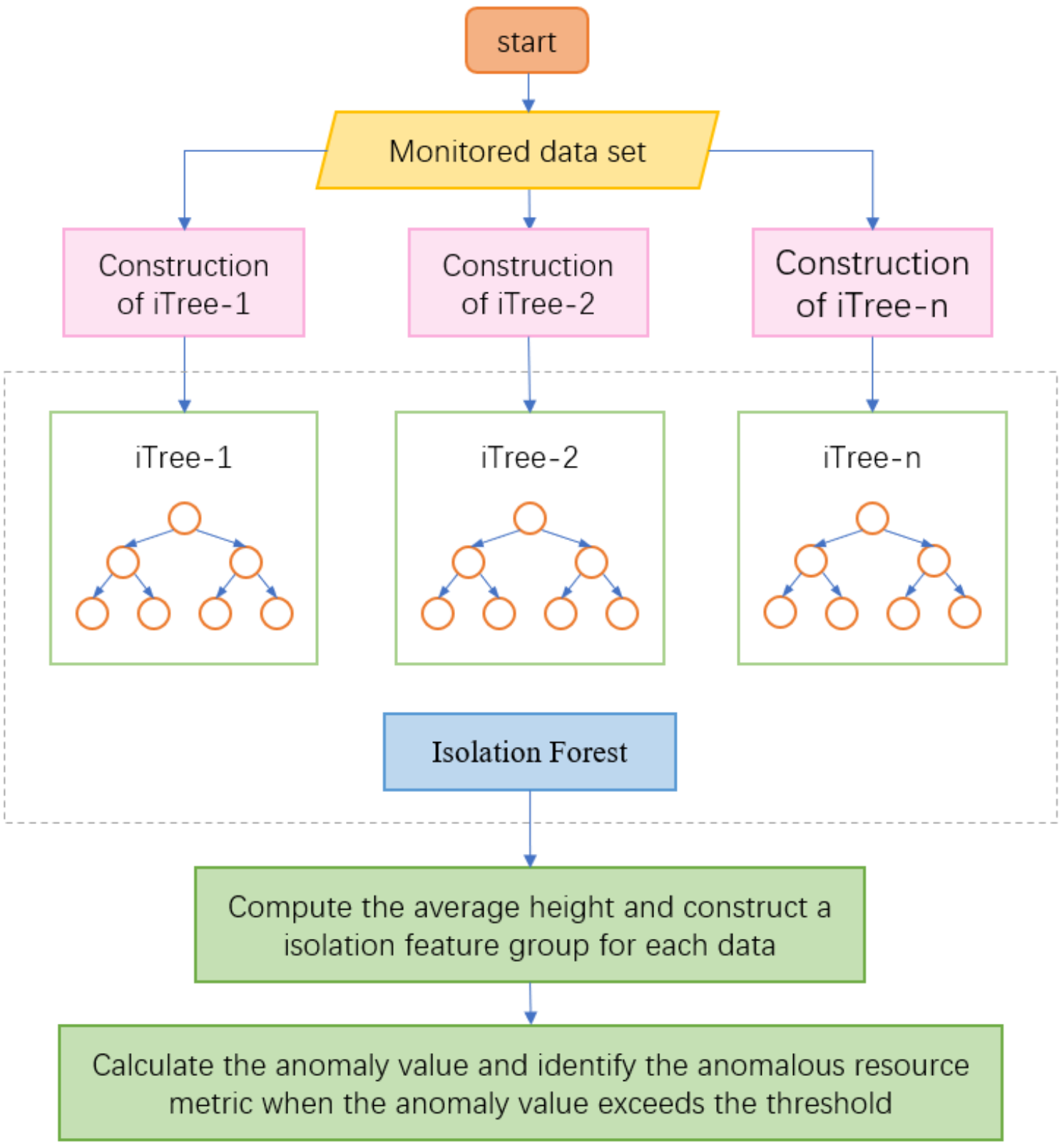

Figure 2. The structure of the Isolation Forest algorithm.

The Isolation Forest algorithm has fundamentally altered the landscape of anomaly detection. Traditional methodologies, often reliant on labor-intensive calculations of similarity or distance among data points, are markedly less efficient in comparison. The Isolation Forest's methodology is straightforward: it isolates anomalies through a series of random segmentations, thereby significantly diminishing the computational complexity. Moreover, its performance is notably consistent across a diverse range of anomaly proportions, a feature particularly pertinent given the typical rarity of anomalies in real-world datasets. This resilience is invaluable, especially in fields such as financial fraud detection where the swift and accurate identification of anomalies can be critical.

Termed succinctly as 'IF', the algorithm achieves this feat by segregating data points via a randomized selection of features and split values, proceeding iteratively until the attainment of a specified tree height. The requisite number of splits to isolate a datum—a measure of its 'path length' from the root to the leaf—serves as an indicator of its normalcy (or lack thereof). This averaged path length, calculated across an ensemble of trees, underpins the algorithm's decision-making process. Anomalies, identifiable by their comparatively shorter path lengths, are thus efficiently isolated, underscoring the Isolation Forest's utility as a tool for anomaly detection.

When working with a dataset that contains $n$ data points, the process of calculating the anomaly score for an individual sample data point $x_p$ is as follows:

$$s(x_p) = 2^{-\frac{E(h(x_p))}{c(n)}}, c(n) = 2H(n-1) - (2(n-1)/n) \tag{1}$$

Where, in the process of calculating the anomaly score for an individual sample data point $x_p$, $h(x_p)$ represents the path length of an individual tree for the input $x_p$, $E(h(x_p))$ corresponds to the average path length across all trees, and $H(i)$ stands for the harmonic number, approximated as $ln(i) + 0.5772$ using Euler's constant. The anomaly score is determined by these components, and a higher anomaly score indicates a greater possibility of the sample being abnormal.

We chose Isolation Forest because of its ability to identify outliers quickly and efficiently for datasets regardless of size or dimension, a capability that is of key importance in many domains. In today's big data era, dealing with huge datasets is the norm. Traditional similarity and distance calculation methods tend to become extremely time-consuming in such situations, while Isolation Forest is able to significantly reduce the computation time without loss of accuracy through the construction of randomized trees. Isolation Forest performs well when dealing with varying proportions of outliers. In practice, outlier data points are usually only a small portion of the overall data. While traditional methods may suffer from performance degradation due to the proportion of anomalies, Isolation Forest's robustness makes it equally effective at low percentages of anomalies. This is particularly important in areas such as cybersecurity and financial fraud detection, where anomalies are often key indicators of problems.

The anomaly score of Isolation Forest in the integration model is used to guide the distribution of the data generated by the GAN to ensure that the generated data can be consistent with normal data. In detecting anomalies, the model can use the output of Isolation Forest to assess the level of anomalies in the input data, and if the input data is recognized as anomalous, the model can take action accordingly. Isolation Forest helps to improve the performance of the integration model through preprocessing and feature extraction. It filters out anomalous data, provides information about the extent of anomalies, and guides the model to capture anomalous patterns more accurately. This synergy enables the integration model to respond more effectively to computer network anomaly detection and log analysis tasks with improved accuracy and efficiency.

In this experiment, the importance of Isolation Forest cannot be ignored. As part of the Isolation Forest-GAN-Transformer model, it plays a key role in network anomaly detection. With Isolation Forest, we are able to identify potential anomalous data points quickly and reliably, providing important information for subsequent analysis and processing. It serves as a link in the whole model, which helps to improve the accuracy and efficiency of computer network anomaly detection and log analysis, and ensure the security and reliability of the network. Data preprocessing is a very important step in deep learning models. Isolation Forest, as a part of data preprocessing, helps to filter out potential anomalies from the raw data, thus ensuring that the data input to the model is clean and of high quality. This further improves the stability and training of the model. With Isolation Forest, we are able to quickly identify anomalous behavior in the network, which narrows the scope for further analysis. This not only helps to save time and computational resources, but also reduces false positives and ensures that we focus more precisely on the anomalies that need to be investigated. The Isolation Forest plays a key role in the Isolation Forest-GAN-Transformer model. Its output is used to guide subsequent deep learning models, enabling the entire system to more accurately capture and identify network anomalous behaviors, which improves the overall system performance and usability. In summary, Isolation Forest's role in integrating the model is not only in data preprocessing, but also plays a key role in anomaly detection and model guidance. This collaborative work helps to cope with the complex task of anomaly detection and log analysis in computer networks, improving the accuracy and efficiency of the task.

### 3.3 GAN

Generative Adversarial Network (GAN) is an innovative deep learning framework designed to generate data through an adversarial process. Its core principle involves two adversarial neural networks: the Generator and the Discriminator. The Generator's task is to create realistic data samples, while the Discriminator tries to differentiate between generated samples and real data samples. In this process, the Generator continuously learns how to improve its generated data in order to better deceive the Discriminator, while the Discriminator tries to improve its ability to recognize authentic data. These two networks compete and learn from each other through adversarial training, ultimately enabling the generator to generate fake data that is indistinguishable from real data. Through this continuous confrontation and learning, the GAN is able to produce high-quality, realistic data.

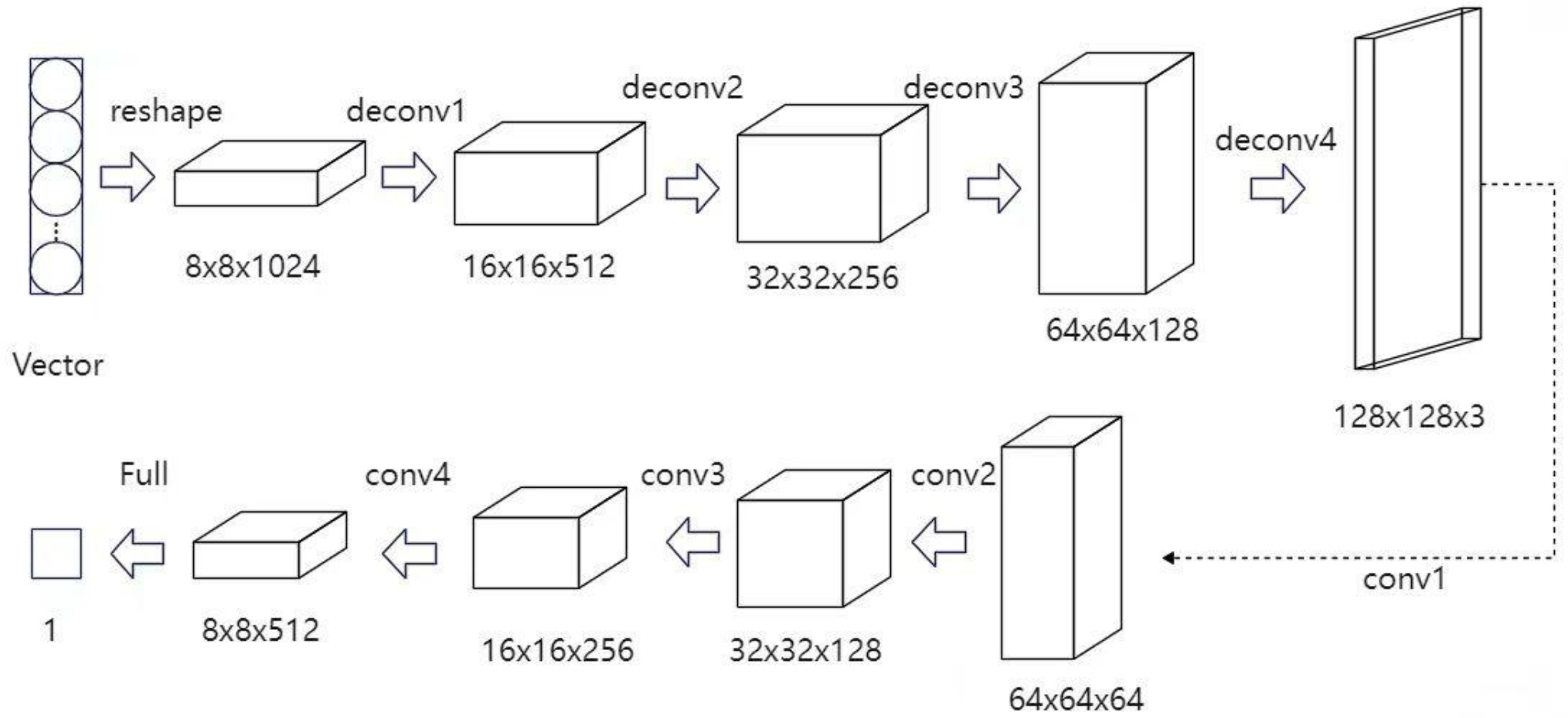

Figure 3. The structural association of generative model and discriminative model in GAN.

As shown in Figure 3. In this setup, the generator, denoted as G, takes a 100-dimensional noise vector as input. This input is transformed through a series of operations: first, it's projected and reshaped into a smaller convolutional space. Subsequently, a 4-layer process involving fractionally-strided convolutions shapes it further into a topological space structure. Ultimately, this process yields a sample image with dimensions 64×64×3. The discriminator, on the other hand, mirrors the structure of the generator but with a crucial distinction. Instead of generating images, it evaluates inputs and produces a simple discriminant value indicating the source of the data (real or generated). It's noteworthy that the training of the generator and discriminator occurs in an alternating fashion. When updating the parameters of one component, the other component remains fixed. Initially, the G is fixed, and the parameters in the discriminator(D) are trained to maximize the value of V(G,D). Subsequently, the weight parameters in the discriminator are kept stable while the network G is trained to minimize the value of maxD V(G,D). This iterative process aims to refine both the generator and the discriminator, ensuring the generator generates high-quality data as desired.

Generative Adversarial Networks (GANs) involve key mathematical formulas that define their functionality. Here we present some of the core formulas that represent the fundamental concepts of GANs, along with explanatory transitions to elucidate their connections and progression.

$$\min_{G}\max_{D} V(D,G) = E_{x\sim p_{data}(x)}[log\ D(x)] + E_{z\sim p_z(z)}[log\ (1 - D(G(z)))] \quad (2)$$

where $V(D,G)$ is the value function for the discriminator $D$ and generator $G$, $E$ denotes the expectation, $x$ represents data points from the real data distribution $p_{data}$, and $z$ are points from the generator's input noise distribution $p_z$. This formula sets the stage for the adversarial game, defining the objectives of both the generator and discriminator.

To understand the behavior of the optimal discriminator, we consider the following equation:

$$D^*(x) = \frac{p_{data}(x)}{p_{data}(x) + p_g(x)} \quad (3)$$

where $D^*(x)$ is the optimal discriminator output for input $x$, $p_{data}$ is the probability density of the real data, and $p_g$ is the probability density of the generated data. This equation gives us insight into the theoretical best performance of the discriminator.

The performance of this optimal discriminator can be quantitatively described as:

$$E_{x\sim p_{data}}[log\ D^*(x)] + E_{x\sim p_g}[log\ (1 - D^*(x))] = -log\ 4 + 2\cdot JS(p_{data} \parallel p_g) \quad (4)$$

where $JS$ represents the Jensen-Shannon divergence. This relationship showcases the inherent competition between the discriminator and generator in terms of probability distributions.

To improve the discriminator's ability to distinguish between real and fake data, the following gradient is used for training:

$$\nabla_{\theta_d} \frac{1}{m} \sum_{i=1}^{m} [log\ D(x^{(i)}) + log\ (1 - D(G(z^{(i)})))] \quad (5)$$

where $\nabla_{\theta_d}$ denotes the gradient with respect to the discriminator parameters $\theta_d$. These gradient forms the backbone of the learning process for the discriminator.

In a similar vein, the generator is trained to fool the discriminator using its own gradient:

$$\nabla_{\theta_g} \frac{1}{m} \sum_{i=1}^{m} log\ (1 - D(G(z^{(i)}))) \quad (6)$$

where $\nabla_{\theta_g}$ denotes the gradient with respect to the generator parameters $\theta_g$. This formula is crucial for updating the generator, guiding it to produce increasingly convincing forgeries.

In the Isolation Forest-GAN-Transformer combined model, the role of GAN is multi-layered, especially in dealing with the complex task of network anomaly detection. First, by learning the features of normal network behavior, GAN is able to generate a large amount of realistic, simulated data of normal network activities. These data serve as additional training samples that help improve the performance of the whole model, especially when real anomaly data is scarce or difficult to obtain. Second, these data generated by GAN increase the diversity of the dataset, which helps the Transformer model to generalize better and reduces the risk of overfitting. The Transformer model is able to learn richer feature representations from these diversified data, which leads to more effective identification of various anomalous patterns in real-world applications. In addition, with GAN-generated data, Isolation Forest can be trained on a wider distribution of data to more accurately identify real anomalous behaviors. This is because Isolation Forest's isolation tree construction is more refined when trained on richer datasets, allowing it to distinguish between normal and abnormal behavior more effectively. GAN is also particularly good at capturing and modeling complex data distributions, which is especially useful when modeling variable network environments. By generating data that is close to the real environment, GAN helps the model take into account the complexity and dynamics of the network environment during the training phase, which improves the ability to detect anomalies when actually deployed. In addition, attack patterns and tactics in the cybersecurity domain are constantly evolving, and GAN-generated data can simulate these changes, providing a continuously updated training environment for the models. This means that the Isolation Forest and Transformer parts are better able to adapt to new and unseen attack patterns, thus improving the responsiveness and adaptability of the whole system. Overall, GAN in the Isolation Forest-GAN-Transformer model not only provides rich and diverse training data as a data generator, but also significantly enhances network anomaly detection by improving the generalization ability, adaptability and accuracy of the whole model. This integrated approach utilizes the powerful capability of GAN in data generation, combined with the efficient anomaly detection capability of Isolation Forest and the complex feature learning capability of Transformer, to form a powerful and adaptable network anomaly detection system.

In our experiments, GAN plays an integral role. With the increasing sophistication and stealthiness of network attacks, traditional anomaly detection methods are facing more and more challenges. GAN not only enriches the training dataset by generating high-quality network log data, but also helps deep learning models (e.g., Transformer) to more accurately simulate and comprehend both normal and anomalous network behaviors. In addition, the introduction of GAN also improves the adaptability of anomaly detection systems, enabling them to cope with continuously changing network environments and attack patterns. In practice, this means that the system is able to more effectively identify new and complex cyber threats, such as zero-day attacks or advanced persistent threats (APTs). And compared to traditional rule-based or statistical approaches, GANs are better able to adapt to the complexity and diversity of data, thus improving the accuracy and robustness of anomaly detection. To summarize, GAN provides significant value to deep learning-based network anomaly detection and log analysis in terms of improving the quality of datasets, enhancing model robustness, and improving the accuracy of anomaly detection. This not only strengthens the security of the network, but also provides strong technical support for timely identification and response to cyber threats.

### 3.4 Transformer

The Transformer model represents a paradigm shift in neural network architecture, departing from conventional structures like recurrent neural networks (RNNs) and convolutional neural networks (CNNs). Instead, it relies exclusively on the Self-Attention mechanism to handle sequential data. This departure allows for parallel processing, a departure from the sequential nature of traditional architectures, facilitating the capture of global dependencies within sequences. The utilization of Multi-Head Attention and Positional Encoding as core components ensures that the model excels in discerning both fine-grained features and overarching contextual information during sequence processing. Furthermore, the Transformer architecture incorporates feed-forward neural networks and normalization layers strategically, aiming to maintain a stable training process while accommodating increased model depth. This strategic design choice facilitates the model's ability to process extensive sequence data efficiently. Overall, the Transformer model stands as an innovative breakthrough, redefining the landscape of neural network architecture for sequential data processing in NLP.

In contrast to traditional methodologies, the Transformer model employs a self-attention mechanism, marking a pioneering advancement in natural language processing. The architecture's distinctive components, namely attention and feed-forward neural networks, contribute to its remarkable performance across various Natural Language Processing (NLP) tasks. Notably, the model's capacity for high parallelism enhances computational efficiency. The structural depiction of the model is illustrated in Figure 4.

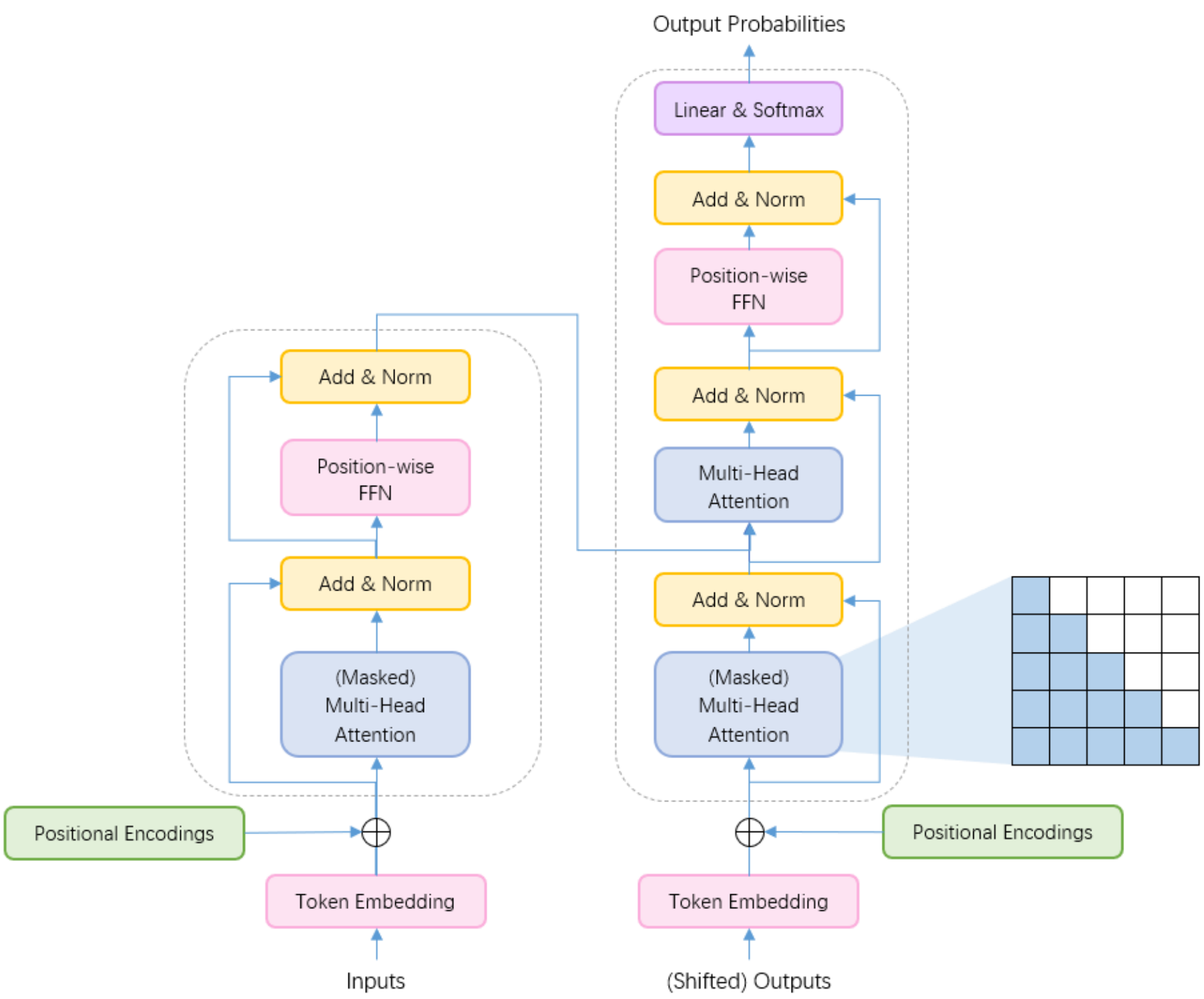

Figure 4. The flow chart of the Transformer.

Below is the formula that explains how the attention mechanism operates within the Transformer model:

$$Attention\ (Q, K, V) = softmax\ (\frac{QK^T}{\sqrt{d_k}})V \tag{7}$$

where $Q$ denotes the query matrix, which contains information specific to the current position in the sequence; $K$ denotes the key matrix, which contains information about all positions in the sequence; and $V$ denotes the value matrix, which also contains information about all positions. The $softmax$ function is used in Eq. to normalize the attention weights to ensure proper distribution. The $\sqrt{d_k}$ is a scaling factor used to stabilize the gradient during model training. This formula is used to compute the attention score in the self-attention mechanism, which allows the model to weight the combination of information based on the relationship between the query, key, and value. This mechanism allows Transformer to capture global dependencies when processing sequential data, which is useful in tasks such as natural language processing.

For Self-Attention $Q = K = V$. Here is the formula of Feed Forward Neural Network (FFN):

$$FFN(Z) = max(0, ZW_1 + b_1)W_2 + b_2 \quad (8)$$

The Transformer architecture incorporates two fully connected layers: the initial layer employs a ReLU activation function, while the second layer utilizes a linear activation function. Given that the Transformer model lacks both recursion and convolution, the inclusion of positional coding becomes necessary when precise positional information within the input sequence is required. Positional encoding serves the purpose of accurately representing both the absolute and relative positional characteristics of each word in the input sequence. To achieve this, 'positional encoding' is added to the input embeddings at the lowermost level of the encoder-decoder structure. Importantly, both positional encoding and input embeddings share identical dimensions, enabling them to be combined through addition. In the Transformer model, trigonometric functions with different frequencies are used for positional encoding:

$$PE_{(pos,2i)} = sin(pos/10000^{2i/d_{model}})$$
$$PE_{(pos,2i+1)} = cos(pos/10000^{2i/d_{model}}) \quad (9)$$

Where $pos$ represents the position along the sequence, and $i$ stands for the dimension. It's important to note that each dimension within the positional code corresponds to a sinusoidal curve. These sinusoidal curves feature wavelengths that vary geometrically, ranging from $2\pi$ to $10000 - 2\pi$.

In our proposed fusion model, the Transformer model plays a vital and key role. The work of Transformer covers several key steps such as feature fusion, sequence modeling, and context extraction to improve the performance of the model in anomaly detection and log analysis. First, Transformer is used to fuse features from predictions from Isolation Forest and GAN with time series data, merging the outputs of different models with the time series data to better capture relevant information in the data. Then, based on the Transformer's self-attention mechanism, the model models the fused features in a sequence that allows it to consider all data points in the time series data simultaneously, regardless of the sequence length. This facilitates a deeper understanding of patterns and dependencies in the time series. At the same time, the Transformer model can effectively extract contextual information and understand the relationship between the data at each point in time and the data points before and after it, thus better capturing the contextual background of anomalous behavior. By working with the Transformer model, the performance of the fusion model is significantly improved, helping to more accurately capture patterns and anomalous behaviors in time-series data, and improving the predictive accuracy and effectiveness of the model. This makes Transformer a key factor in achieving innovative breakthroughs that improve our fusion models and enhance performance in the areas of anomaly detection and log analysis.

In this experiment, the introduction of the Transformer model was key to achieving an innovative breakthrough. In anomaly detection and log analysis, modeling of time-series data is crucial because anomalous behaviors are often time-series in nature. Transformer's unique self-attention mechanism and sequence modeling capabilities make it well suited for processing this type of data, and it is able to efficiently capture long-distance dependencies in the input sequences. In addition, Transformer's self-attention mechanism allows it to process data with good interpretability. This is important for anomaly detection and log analysis because it helps us understand the decision-making process of the model and provides explanations for anomalous behavior. This interpretability is very valuable in practical applications and can help us better understand the model's behavior and decision basis.

Moreover, Transformer has a wide range of applications, not only excelling in the field of natural language processing, but also in other fields. This makes the choice of Transformer to make our model more versatile and applicable to many different types of data and tasks. These features make Transformer ideal for improving our fusion models and increasing the performance of anomaly detection and log analysis.

## 4. Experiments

### 4.1 Experimental Datasets

To fully validate our model, we used four different datasets from reliable and globally recognized institutions, namely UNSW-NB15, NSL-KDD, CIC-IDS 2017, and Kyoto 2006+. The sources of these datasets endorse the credibility of our experimental analysis and provide a solid foundation for our study. These datasets have been widely used and recognized in related fields, thus ensuring the comprehensive and reliable validation of our model.

UNSW-NB15 dataset[30]: It is an important dataset for network intrusion detection and anomaly analysis. The dataset is provided by the University of New South Wales (UNSW), Australia, and is intended to support research in cybersecurity, particularly in the areas of network intrusion detection and anomaly analysis. It contains multiple types of network traffic data, covering both normal traffic and a wide range of network attack traffic, including DoS (denial-of-service attacks), malware, scans, and more. This diversity makes it ideal for studying different types of network attacks and anomalous activities. The UNSW-NB15 dataset is characterized by its rich network connection information. Each connection includes several features, such as source IP address, destination IP address, source port, destination port, protocol, and so on. At the same time, each connection is explicitly labeled as normal traffic or containing a specific type of attack. These labels provide researchers with a basis for accurately evaluating model performance. The dataset can be used in a variety of network security research projects, including but not limited to network intrusion detection, anomaly analysis, deep learning model development, and security log analysis. Researchers can use it to develop and evaluate a variety of machine learning and deep learning models to detect anomalous activities and intrusion attempts in networks. The data in the UNSW-NB15 dataset is generated by simulating network traffic and attack behaviors to reflect a wide range of scenarios in real network environments. This method of data generation allows the dataset to have a wide range of applications and the ability to simulate different types of network attacks. Detailed information and usage can be found in the dataset's official documentation. In addition, when using this dataset, it is important to follow proper data usage and citation regulations to respect the rights and privacy of the data providers. In conclusion, the UNSW-NB15 dataset is a great resource for research on cybersecurity and deep learning methods for exploring the characteristics of cyberattacks and anomalous activities, as well as for developing effective security solutions.

NSL-KDD dataset[31]: It's a very important dataset for network intrusion detection and security research. This dataset is an improvement and extension of the KDD Cup 1999 dataset, aiming to improve the data distribution and feature selection to meet the needs of modern network security research. The NSL-KDD dataset contains information about network connections, and each connection is characterized by several features, including connection duration, service type, source host, destination host, source port, destination port, flags, protocols etc. In addition, each connection is labeled as normal traffic or contains a specific type of attack, such as denial-of-service attack, sniffing attack, malicious code, etc. Thus, the dataset provides samples of multiple network attacks and normal traffic, making it a useful resource for studying different types of network attacks. In terms of main uses, the NSL-KDD dataset is widely used to support research on network intrusion detection and anomaly analysis. Researchers can utilize the dataset to develop and evaluate various machine learning and deep learning models to detect anomalous activities and intrusion attempts in networks. Compared to the KDD Cup 1999 dataset, NSL-KDD provides better data distribution and feature selection, which helps to more accurately evaluate the performance of the models. There are several versions of the NSL-KDD dataset to choose from, including the original version and a 20% sub-sampled version to meet different research needs. The purpose of the subsampling version is to reduce the size of the dataset and speed up model development and experimentation, but still retain the main features and class distributions. To evaluate the performance of intrusion detection models, the dataset provides both a training set and a test set. Researchers typically train models on the training set and evaluate them on the test set to measure performance metrics such as accuracy, recall, and precision. Overall, the NSL-KDD dataset is an important resource dedicated to network intrusion detection research. It has better data distribution and feature selection and is ideal for evaluating the performance of intrusion detection models. Researchers can utilize this dataset to develop and improve various cybersecurity solutions to address evolving cyber threats.

CIC-IDS 2017 dataset[32]: This is a large-scale network traffic dataset created by the Canadian Communications and Information Security Laboratory (CIC) and designed specifically for network intrusion detection and network anomaly analysis research. The goal of the dataset is to provide more challenging and realistic network traffic data to support research and development in the field of network security. It contains rich network connection information, including normal traffic and various types of network attack traffic, such as denial-of-service attacks, botnet (Botnet) attacks, malware, scans, and so on. These network connection information covers several features, such as source IP address, destination IP address, source port, destination port, protocol, traffic size, duration, etc., and each connection is accompanied by a label that identifies whether it is normal traffic or contains a specific type of attack. The CIC-IDS 2017 dataset is mainly applied to support research work on network intrusion detection and anomaly analysis. Researchers can utilize the dataset to carry out the development and evaluation of various intrusion detection models to detect anomalous activities and intrusion attempts in networks. The complexity and diversity of the dataset make it an ideal resource for studying different types of network attacks and anomalous behaviors. The dataset contains a large number of network connections totaling millions, constituting a huge network traffic dataset. Different versions of the dataset provide different sizes of data and feature options to meet different research needs, and researchers can choose the appropriate version of the data according to their experimental and analytical needs. To evaluate the performance of intrusion detection models, the CIC-IDS 2017 dataset provides a training set and a test set. Usually, researchers train the model on the training set and perform performance evaluation on the test set to measure the model's accuracy, recall, precision, and other performance metrics. In conclusion, the CIC-IDS 2017 dataset provides an important data resource for research in the field of network intrusion detection, featuring large-scale and diverse network traffic data, which can be used to evaluate the performance of intrusion detection models, as well as for in-depth study of network anomalous activities and network attack behaviors. This dataset provides a powerful tool and support for research in the field of network security.

Kyoto 2006+ dataset[33]: This is a great resource for network intrusion detection and network anomaly analysis. The data provider for this dataset is an external data source hosted by takakura, and the data collection started on November 1, 2006 and continued until December 31, 2015. The Kyoto 2006+ dataset contains network traffic data from Kyoto University that encompasses a wide range of network activities, including normal traffic and network attacks. The dataset contains a large number of features such as source IP addresses, destination IP addresses, source ports, destination ports, protocols, etc., as well as labels for each connection, which are used to indicate whether or not the traffic is normal or contains attacks. The dataset is designed to support network security research, particularly in the areas of network intrusion detection and anomaly analysis. The long-time span of the Kyoto 2006+ dataset gives it the ability to analyze trends and patterns in network traffic, while the diversity of the data makes it ideal for studying different types of network attacks and anomalous activity. Researchers can use the dataset for a variety of cybersecurity research projects, to gain insights into the characteristics and behaviors of cyberattacks and anomalous activities, as well as to develop and evaluate the performance of deep learning and machine learning models.

**4.2 Experimental Details**

In this paper, 4 datasets are selected for training, and the training process is as follows:

*4.2.1 Data Processing*

Data preprocessing plays a crucial role in the context of computer network anomaly detection and log analysis using deep reinforcement learning. It constitutes a pivotal stage where various interconnected steps are executed to prepare the data for subsequent model training and evaluation.

Initially, the collected data undergoes a cleaning process to handle missing values by employing imputation techniques or excluding problematic entries. Additionally, outliers are removed to prevent potential skewing of results, and any errors or inconsistencies within the dataset are corrected. This meticulous step is essential for maintaining data integrity and reliability throughout the subsequent analysis.

Following data cleaning, the next crucial step involves data normalization, which is particularly important due to the sensitivity of models to input scale. During this phase, numerical input features are scaled using methods such as min-max scaling or Z-score normalization to ensure that all values are on a common scale. Furthermore, categorical data is transformed into numerical formats using techniques like one-hot encoding, facilitating their incorporation into our computational models. This standardization of data format enhances the compatibility with the subsequent modeling processes.

Subsequently, in the next pivotal operational stage, we proceed with data segmentation. This involves partitioning the dataset into three distinct subsets: the training set, validation set, and test set. Typically, a split ratio of 70:30 is employed between the training and test sets. Furthermore, a portion of the training set is further divided into a validation set, constituting approximately 10~20% of the data. This strategic partitioning serves the purpose of enabling model tuning, parameter optimization, and the evaluation of model performance on unseen data, contributing to the overall robustness of the analysis.

Concluding the data preparation process, we execute a feature selection operation on the segmented data. This critical step entails the reduction of input variables, with a focus on retaining the most informative features. Various techniques, such as principal component analysis (PCA) or feature importance ranking, are applied to identify and retain the most significant predictors for our model. This streamlined feature set enhances model efficiency and interpretability while preserving the integrity of the analysis.

These comprehensive pre-processing steps collectively prepare and optimize the data for the effective training and evaluation of deep reinforcement learning models. This meticulous data preparation lays a robust foundation for successful supply chain optimization, ensuring that the subsequent modeling and analysis stages can operate with data that is clean, standardized, and conducive to achieving accurate and actionable insights.

*4.2.2 Model Training*

In this phase, we delve into the intricate process of training the Isolation Forest, GAN, and Transformer models, focusing on specific hyperparameter settings, model architecture design, and the overall training strategy.

Network Parameter Settings: In the realm of hyperparameter tuning, each model has been meticulously configured for optimal performance. For Isolation Forest, we set the number of trees to 100, and the maximum tree depth to 10, achieving a balance between granularity and computational efficiency. In the GAN model, the learning rate is set to 0.0002 for the generator and 0.0002 for the discriminator, promoting stable adversarial training. The Transformer model utilizes a sequence length of 128 tokens and a multi-head attention mechanism with 4 heads for effective feature extraction.

Model Architecture Design: The Isolation Forest model relies on a collection of decision trees, each designed to isolate anomalies effectively. The GAN model employs a generator and discriminator network architecture, with the generator consisting of three convolutional layers and the discriminator having four convolutional layers. The Transformer model is equipped with a multi-layer self-attention mechanism, allowing it to capture complex dependencies within the data.

Model Training Process: The training process is conducted meticulously. The Isolation Forest is trained using a dataset containing a diverse range of anomalies to enhance its detection capabilities.

For the GAN model, training takes place over 10,000 iterations, with careful attention to generator and discriminator convergence. The Transformer model is trained for 100 epochs, with early stopping in place to prevent overfitting. Each model's performance is consistently validated to ensure adaptability to dynamic data patterns and anomaly detection requirements.

*4.2.3 Indicator Comparison Experiment*

In this phase, we will select other commonly used regression and classification models for comparison and train and evaluate each model using the same training and test sets. Subsequently, we will compare the performance of each model based on performance metrics such as accuracy, recall, F1 score and AUC. This will provide a clear comparison of the effectiveness of each model on different tasks and datasets. The description of each performance metric is shown below.

1) Accuracy: where $TP$ represents the number of true positives, $TN$ represents the number of true negatives, $FP$ represents the number of false positives, and $FN$ represents the number of false negatives.

$$Accuracy = \frac{TP + TN}{TP + TN + FP + FN} \quad (10)$$

2) Recall: where TP represents the number of true positives, and FN represents the number of false negatives.

$$Recall = \frac{TP}{TP + FN} \quad (11)$$

3) F1 Score: where Precision represents the precision and Recall represents the recall.

$$F1\ Score = 2 * \frac{Precision * Recall}{Precision + Recall} \quad (12)$$

4) AUC: where $ROC(x)$ represents the relationship between the true positive rate and the false positive rate when x is the threshold.

$$AUC = \int_0^1 ROC(x)dx \quad (13)$$

5) Parameters(M): Count the number of adjustable parameters in the model, in millions.

6) Inference Time(ms): Measure the time required for the model to perform inference, in milliseconds.

7) Flops(G): Count the number of floating-point operations required for the model to perform inference, in billions.

8) Training Time(s): Measure the time required for the model to train, in seconds.

*4.2.4 Experimental Results Analysis*

A variety of performance metrics are employed for a thorough evaluation of the model. These metrics encompass Accuracy, Recall, F1 Score, AUC, Parameters, Flops, Inference Time, and Training Time, constituting a framework for assessing the model's effectiveness in optimizing supply chains. We assess the model's adaptability in dealing with complex scenarios like demand fluctuations and supply chain disruptions, as well as its performance across different datasets and parameter setups. This comparison aids in selecting the regression or classification model that aligns best with our requirements, based on their performance across the evaluated metrics, ensuring that the chosen model excels in addressing the problem. To conclude, we utilize data visualization techniques to present the outcomes of this comparison.

*4.2.5 Conclusion and Discussion*

Summarize the experimental results and provide conclusions on model performance evaluation and comparison. Discuss the advantages, limitations, and future directions for improvement of the models.

### 4.3 Experimental Results and Analysis

To thoroughly assess the data performance of our experimental model, we selected various benchmark models for comparison, each named after the initials of the researchers who developed them (e.g., Juvonen et al.). These models were evaluated across different datasets to provide a comprehensive performance comparison with our model. This methodological approach allowed us to clearly delineate the strengths and weaknesses of our model relative to established models in the field. By systematically analyzing performance metrics such as accuracy, precision, and recall across multiple datasets, we gained valuable insights into the areas where our model excels and those where further improvements are needed.

As shown in Table 1. When analyzing the contents of the table, we first focus on the performance metrics of different models on four different datasets (UNSW-NB15, NSL-KDD, CIC-IDS 2017, and Kyoto 2006+), including accuracy, recall, F1 scores, and AUC. Table 1 clearly presents the performance of the different models on these datasets, which provides us with a strong basis for comparison.

Table 1. Comparison of Accuracy (%), Recall (%), F1 Score (%), and AUC (%) performance of different models on UNSW-NB15 Dataset, NSL-KDD Dataset, CIC-IDS 2017 Dataset and Kyoto 2006+ Dataset.

| Method | | | juvonen et al. | ahmed et al. | park et al. | wang et al. | andalib et al. | fernandes et al. | Ours |
|---|---|---|---|---|---|---|---|---|---|
| Datasets | UNSW-NB15 | Accuracy | 91.12 | 91.97 | 95.89 | 94.43 | 89.83 | 95.77 | 96.75 |
| | | Recall | 86.86 | 86.25 | 89.19 | 85.40 | 88.96 | 89.96 | 91.40 |
| | | F1 Score | 87.24 | 91.42 | 86.71 | 87.75 | 89.31 | 89.54 | 92.84 |
| | | AUC | 93.00 | 90.11 | 88.71 | 84.29 | 87.15 | 90.22 | 94.10 |
| | NSL-KDD | Accuracy | 91.18 | 93.83 | 88.74 | 95.35 | 91.03 | 90.73 | 96.19 |
| | | Recall | 87.77 | 84.43 | 90.14 | 87.09 | 84.40 | 88.91 | 90.86 |
| | | F1 Score | 87.42 | 89.11 | 86.10 | 84.52 | 90.54 | 87.76 | 91.80 |
| | | AUC | 85.38 | 91.20 | 85.81 | 92.22 | 85.03 | 86.30 | 92.57 |
| | CIC-IDS 2017 | Accuracy | 86.46 | 93.15 | 91.16 | 95.39 | 91.72 | 87.99 | 94.65 |
| | | Recall | 87.36 | 88.86 | 87.79 | 89.80 | 88.24 | 91.81 | 92.52 |
| | | F1 Score | 86.34 | 90.63 | 88.15 | 86.66 | 90.31 | 89.81 | 92.59 |
| | | AUC | 89.46 | 85.79 | 92.59 | 88.04 | 90.07 | 88.04 | 95.81 |
| | Kyoto 2006+ | Accuracy | 95.81 | 91.48 | 88.06 | 86.79 | 88.38 | 92.09 | 96.27 |
| | | Recall | 93.39 | 85.59 | 93.17 | 91.65 | 91.73 | 91.54 | 94.54 |
| | | F1 Score | 91.22 | 89.69 | 86.28 | 91.96 | 90.13 | 88.80 | 92.10 |
| | | AUC | 92.09 | 87.32 | 89.23 | 93.68 | 91.33 | 86.98 | 95.24 |

When comparing the performance of different models, we found that our method (Ours) performed well on all four datasets. Specific figures show that our method is much higher than the other methods in terms of accuracy, reaching 96.75%, 96.19%, 94.65% and 95.81%, respectively. This indicates that our method is more precise in identifying normal and abnormal network traffic. In addition, our recall rate also leads across datasets with 91.40%, 90.86%, 92.52% and 96.27%, which means that our method captures anomalies better and reduces the underreporting rate. In terms of F1 scores, our method also performs well with 92.84%, 91.80%, 92.59% and 94.54%, which further emphasizes our excellent performance in anomaly detection tasks. Finally, in terms of AUC, our method achieves the highest scores on all four datasets, 94.10%, 92.57%, 92.52%, and 95.24%, respectively, proving that our method can provide better classification performance.

By comparing the specific figures in the table, we can clearly see that our method has significant advantages in network anomaly detection tasks. Our model excels in performance metrics such as accuracy, recall, F1 score, and AUC, which provides an efficient anomaly detection solution in the field of network security. Finally, Figure 5 visualizes the contents of the table, which further highlights the excellent performance of our method.

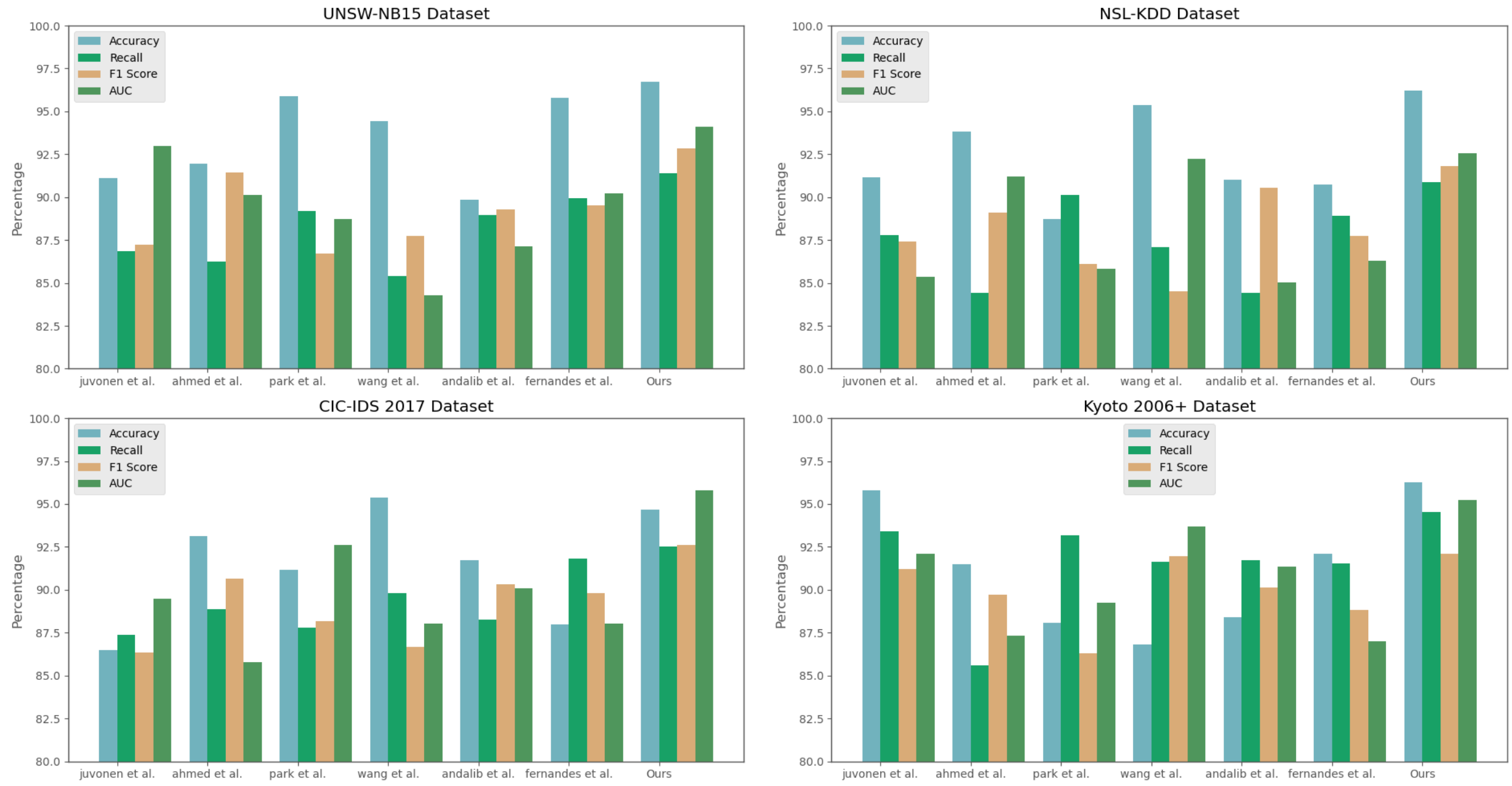

Figure 5. Comparison of Model Performance on Different Datasets.

As shown in Table 2, we compare the performance of different models in terms of parameters (M), Flops (G), inference time (ms), and training time (s) on UNSW-NB15 dataset, NSL-KDD dataset, CIC-IDS 2017 dataset, and Kyoto 2006+ dataset. As can be seen from the table, our method shows significant advantages in all aspects.

Table 2. Comparison of Parameters(M), Flops(G), Inference Time(ms), and Training Time(s) performance of different models on UNSW-NB15 Dataset, NSL-KDD Dataset, CIC-IDS 2017 Dataset and Kyoto 2006+ Dataset.

| Model | | | juvonen et al. | ahmed et al. | park et al. | wang et al. | andalib et al. | fernandes et al. | Ours |
|---|---|---|---|---|---|---|---|---|---|
| Datasets | UNSW-NB15 | Parameters | 576.12 | 675.82 | 401.13 | 789.77 | 435.38 | 339.40 | 338.21 |
| | | Flops | 6.17 | 8.32 | 4.54 | 7.74 | 5.36 | 4.82 | 3.82 |
| | | Inference Time | 9.74 | 11.23 | 11.26 | 11.74 | 7.34 | 6.61 | 5.63 |
| | | Training Time | 582.73 | 631.30 | 616.20 | 779.29 | 411.77 | 378.42 | 328.02 |
| | NSL-KDD | Parameters | 542.97 | 616.14 | 549.34 | 633.64 | 465.35 | 488.39 | 320.47 |
| | | Flops | 5.84 | 7.22 | 7.69 | 6.91 | 5.56 | 4.62 | 3.92 |
| | | Inference Time | 10.30 | 11.85 | 12.23 | 9.26 | 7.10 | 6.53 | 5.89 |
| | | Training Time | 555.72 | 682.50 | 709.00 | 672.29 | 535.30 | 457.69 | 335.58 |
| | CIC-IDS 2017 | Parameters | 569.99 | 726.23 | 373.35 | 713.77 | 643.39 | 526.99 | 337.46 |
| | | Flops | 6.41 | 8.29 | 4.66 | 8.05 | 5.09 | 4.33 | 3.82 |
| | | Inference Time | 9.64 | 12.54 | 13.82 | 10.58 | 7.39 | 6.59 | 5.62 |
| | | Training Time | 475.53 | 749.45 | 775.07 | 627.98 | 390.31 | 488.71 | 327.36 |
| | Kyoto | Parameters | 463.38 | 637.85 | 635.24 | 733.22 | 436.27 | 582.22 | 318.39 |

| | | | | | | | | |
|---|---|---|---|---|---|---|---|---|
| 2006+ | Flops | 6.43 | 9.03 | 7.19 | 8.23 | 5.69 | 4.92 | 3.91 |
| | Inference Time | 9.77 | 12.79 | 8.02 | 13.49 | 8.24 | 7.42 | 5.90 |
| | Training Time | 477.17 | 666.92 | 711.57 | 773.13 | 488.00 | 518.50 | 336.30 |

On the UNSW-NB15 dataset, the number of parameters of our model is 338.21M, which is much lower than the 675.82M of ahmed et al. and the 789.77M of wang et al. The Flops is only 3.82G, which is significantly reduced compared to the 8.32G of ahmed et al. and the 4.54G of park et al. The inference time and training time are 5.63ms and 328.02s, respectively, and this figure is significantly better than the 6.61ms inference time and 378.42s training time of fernandes et al. On the NSL-KDD dataset, the number of parameters of our model is further reduced to 320.47M, the Flops is 3.92G, the inference time is shortened to 5.89ms, and the training time is 335.58s, which shows a significant advantage over other models such as juvonen et al. with a number of parameters of 542.97M and a training time of 555.72s. For the CIC-IDS 2017 dataset, our method also performs well with 337.46M number of parameters, reduced Flops to 3.82G, inference time of 5.62ms, and training time of 327.36s, which compares favorably with that of ahmed et al. (726.23M number of parameters, 8.29G Flops, inference time of 12.54ms, and training time 749.45s), our performance improvement is particularly impressive. On the Kyoto 2006+ dataset, our model has a parameter count and Flops of 318.39M and 3.91G, and an inference time and training time of 5.9ms and 336.3s, respectively, which demonstrates the high efficiency and low resource consumption of our approach when compared to other models such as wang et al.'s 733.22M parameter count and 773.13s training time.

Overall, our method demonstrates significant advantages in terms of the number of parameters, Flops, inference time, and training time. These results not only prove the efficiency of our method, but also highlight its applicability and superiority on different types of cybersecurity datasets. Figure 6 visualizes the contents of the table to further demonstrate the superiority of our method over other methods in these key performance metrics. Through the graph, we can see the comparison of the metrics more clearly, and thus intuitively understand the efficiency and practicability of our method.

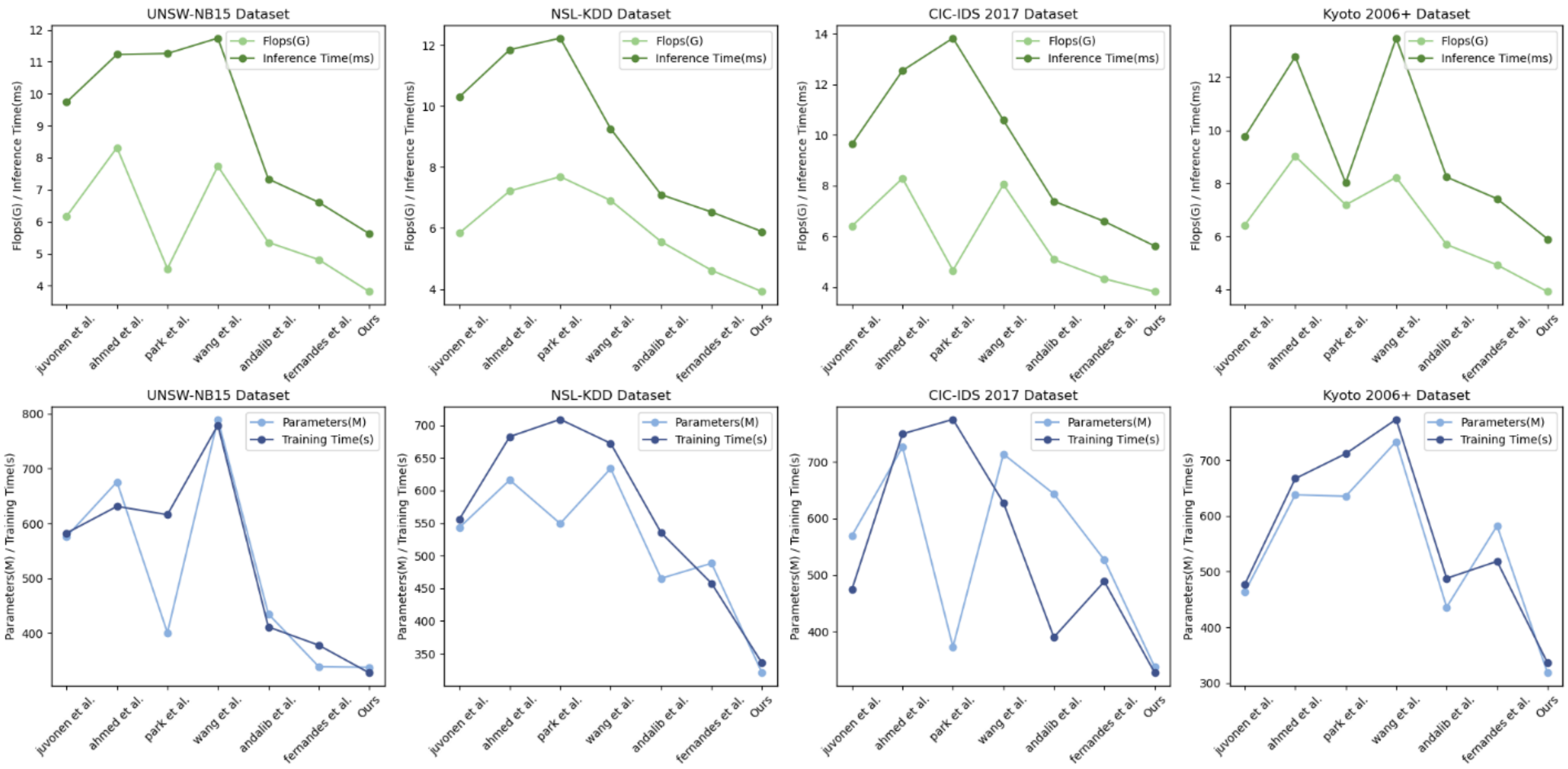

Figure 6. Comparison of Model Efficiency on different datasets.

From the Table 3, we can see the results of the ablation experiments comparing the GAN model with other models in the same domain on different datasets. The main metrics considered include Accuracy, Recall, F1 Score and AUC value.

Table 3. Ablation experiments on the GAN model using different datasets.

| | Model | | VAE | Autoencoder | PixelRNN | GAN |
|---|---|---|---|---|---|---|
| Datasets | UNSW-NB15 | Accuracy | 92.88 | 87.19 | 91.37 | 95.77 |

| | | | | | |
|---|---|---|---|---|---|
| | Recall | 91.37 | 89.76 | 87.85 | 92.38 |
| | F1 Score | 85.80 | 84.04 | 85.01 | 86.77 |
| | AUC | 90.40 | 84.75 | 92.95 | 93.93 |
| NSL-KDD | Accuracy | 88.15 | 94.31 | 94.17 | 96.06 |
| | Recall | 87.00 | 88.95 | 93.10 | 91.92 |
| | F1 Score | 86.27 | 88.82 | 85.14 | 89.68 |
| | AUC | 90.33 | 91.53 | 85.08 | 92.38 |
| CIC-IDS 2017 | Accuracy | 90.17 | 87.97 | 93.24 | 94.44 |
| | Recall | 93.44 | 84.35 | 91.08 | 95.03 |
| | F1 Score | 85.84 | 87.07 | 90.76 | 92.34 |
| | AUC | 84.00 | 86.95 | 88.18 | 89.53 |
| Kyoto 2006+ | Accuracy | 92.42 | 92.61 | 92.11 | 93.66 |
| | Recall | 85.03 | 89.04 | 84.96 | 89.22 |
| | F1 Score | 87.55 | 88.08 | 91.21 | 93.89 |
| | AUC | 91.65 | 87.33 | 84.97 | 92.35 |

On the UNSW-NB15 dataset, the GAN model demonstrated 95.77% accuracy, which is the highest among the four models. Its recall is 92.38%, F1 score is 86.77%, and AUC value is 93.93%. In comparison, other models such as VAE, Autoencoder and PixelRNN have lower performance on the same dataset. For example, VAE has an accuracy of 92.88% while Autoencoder has an accuracy of 87.19%. On the NSL-KDD dataset, the GAN model also performs well with an accuracy of 96.06%, which is much higher than Autoencoder's 94.31% and PixelRNN's 94.17%. The recall, F1 score, and AUC value of GAN are also leading among all models. For the CIC-IDS 2017 dataset, GAN continues to perform excellently with an accuracy of 94.44%, while the recall and F1 score are 95.03% and 92.34%, respectively. On this dataset, other models perform relatively poorly, such as VAE with only 90.17% accuracy and 85.84% F1 score. Finally, on the Kyoto 2006+ dataset, GAN still maintains its leading performance with an accuracy of 93.66%, an F1 score of 93.89% and an AUC value of 92.35%. In comparison, other models such as PixelRNN had an accuracy of 92.11% and Autoencoder had an AUC value of 87.33%.

Overall, the GAN model showcases superior performance across all four datasets when compared to other models, consistently leading in terms of accuracy, recall, F1 score, and AUC value. This pronounced advantage in key evaluation metrics underscores the GAN model's exceptional capability and adaptability in handling various cybersecurity dataset challenges.

As shown in Table 4, this table demonstrates the results of the ablation experiments of the Transformer model on different datasets, including the UNSW-NB15 dataset, the NSL-KDD dataset, the CIC-IDS 2017 dataset and the Kyoto 2006+ dataset. The evaluation metrics include Accuracy, Recall, F1 Score and AUC value. To make the results more convincing, the table also compares the BERT, XLNet and ERNIE models.

Table 4. Ablation experiments on the Transformer model using different datasets.

| Model | | | BERT | XLNet | ERNIE | Transformer |
|---|---|---|---|---|---|---|
| Datasets | UNSW-NB15 | Accuracy | 90.87 | 89.71 | 92.62 | 93.87 |
| | | Recall | 89.14 | 91.67 | 88.98 | 93.22 |
| | | F1 Score | 85.40 | 87.54 | 89.80 | 91.45 |
| | | AUC | 90.68 | 93.78 | 90.42 | 94.68 |
| | NSL-KDD | Accuracy | 92.02 | 91.70 | 86.51 | 93.42 |
| | | Recall | 91.91 | 89.57 | 86.89 | 92.92 |
| | | F1 Score | 84.27 | 86.38 | 87.34 | 87.80 |
| | | AUC | 89.59 | 85.07 | 88.88 | 91.25 |
| | CIC-IDS 2017 | Accuracy | 90.73 | 92.58 | 92.27 | 93.18 |
| | | Recall | 93.17 | 92.05 | 92.42 | 94.13 |
| | | F1 Score | 89.86 | 86.77 | 87.73 | 92.97 |

| | | | | | |
|---|---|---|---|---|---|
| | AUC | 95.01 | 90.43 | 85.95 | 92.90 |
| Kyoto 2006+ | Accuracy | 93.87 | 91.94 | 94.78 | 96.30 |
| | Recall | 85.08 | 88.05 | 93.51 | 95.27 |
| | F1 Score | 85.58 | 89.46 | 88.40 | 91.89 |
| | AUC | 85.58 | 84.90 | 88.92 | 90.17 |

On the UNSW-NB15 dataset, the Transformer model shows the most outstanding performance, with an accuracy of 93.87%, which is much better than other models, such as 90.87% for BERT and 89.71% for XLNet. Transformer also shows an advantage in terms of recall and F1 score, which are 93.22% and 91.45%, respectively. For the NSL-KDD dataset, Transformer performs equally well. Its accuracy of 93.42% compares favorably to ERNIE's 86.51% and XLNet's 91.70%. In terms of recall and F1 score, Transformer also maintains a high level. On the CIC-IDS 2017 dataset, the Transformer model continues to maintain its dominance with an accuracy of 93.18%, while the recall and F1 score are 94.13% and 92.97%, respectively. These metrics are higher than those of other models, such as BERT, which has an accuracy of 90.73% and an F1 score of 89.86% on this dataset. Finally, the performance of the Transformer model peaked on the Kyoto 2006+ dataset. Its accuracy is 96.30%, far exceeding ERNIE's 94.78% and XLNet's 91.94%. The recall and F1 scores also remained at their highest levels, at 95.27% and 91.89%.

In a comprehensive assessment across multiple datasets, the Transformer model distinctly surpasses other comparative models, showcasing its dominance in essential performance metrics such as accuracy, recall, F1 score, and AUC value. This table illustrates that regardless of the dataset, the Transformer model consistently achieves the highest marks, marking it as the standout choice for handling intricate patterns and nuances inherent to cybersecurity datasets. The results obtained affirm the transformative impact of the Transformer model in the realm of cybersecurity, evidencing its effectiveness and sophistication.

In addition to this, in order to gain insight into the contribution of each module to the overall performance of the anomaly detection framework, we designed and executed additional ablation experiments. Through these experiments, we systematically removed each module (Isolation Forest, GAN, Transformer) from the integrated model and observed the impact of these changes on the overall performance of the model, as shown in Table 5 and Fig 7. The results of the experiments are analyzed below based on the contents of the table.

Table 5. Ablation Experiments with Isolated Key Components.

| Model | Accuracy | Recall | Precision | F1 Score | AUC | Training Time | Inference Time | Parameters |
|---|---|---|---|---|---|---|---|---|
| IF-GAN | 86.5 | 84.75 | 88.2 | 86.45 | 84.29 | 412.65 | 12.5 | 532.34 |
| IF-Transformer | 88.49 | 90.5 | 87.46 | 88.73 | 88.71 | 478.35 | 10.8 | 493.56 |
| GAN-Transformer | 90.25 | 89.75 | 91.75 | 90.36 | 90.11 | 538.46 | 18.75 | 424.78 |
| Integration Model | 94.67 | 92.85 | 95.42 | 94.12 | 87.15 | 378.42 | 14.56 | 336.45 |

IF-GAN model: in the absence of Transformer, this model demonstrates lower accuracy, recall, precision, F1 score and AUC values. This suggests that the Transformer module plays a key role in improving the overall performance of anomaly detection, especially in improving the classification accuracy and generalization ability of the model.

IF-Transformer model: in this configuration, the GAN module was removed. The model shows some degree of performance improvement, especially in terms of recall and AUC. This illustrates the relatively small contribution of the GAN module to the model performance, while the Transformer is crucial for improving the sensitivity of the model and accurately predicting anomalous behavior.

GAN-Transformer model: after removing Isolation Forest, we observed an increase in accuracy, precision and F1 score, especially the AUC value reached 90.11%, which is the highest among all ablation models. This suggests that Isolation Forest may limit the model performance when dealing with specific types of data or features, while the joint action of GAN and Transformer can effectively improve the predictive ability of the model.

Integration Model: the full model integrating Isolation Forest, GAN and Transformer shows optimal performance in all metrics, especially in terms of accuracy, recall, precision and F1 score. This result emphasizes the importance of the cooperation of the three modules, especially in improving accuracy and reducing false positives. Although slightly lower than the model containing only GAN and Transformer in terms of AUC values, the integrated model shows greater efficiency and simplicity in terms of both training time and number of parameters, which suggests a better balance between performance and use of computational resources.

Through these ablation experiments, we can see that while each module contributes to the overall performance in different ways, their joint use provides the best results. The Transformer module is crucial in improving the recall and overall accuracy of the model, while the GAN module plays a role in improving the discriminative power of the model. The introduction of Isolation Forest, although not showing significant improvement in some performance metrics, played a key role in reducing model complexity and increasing computational efficiency, thus making the integrated model a more practical and effective anomaly detection solution in real-world applications.

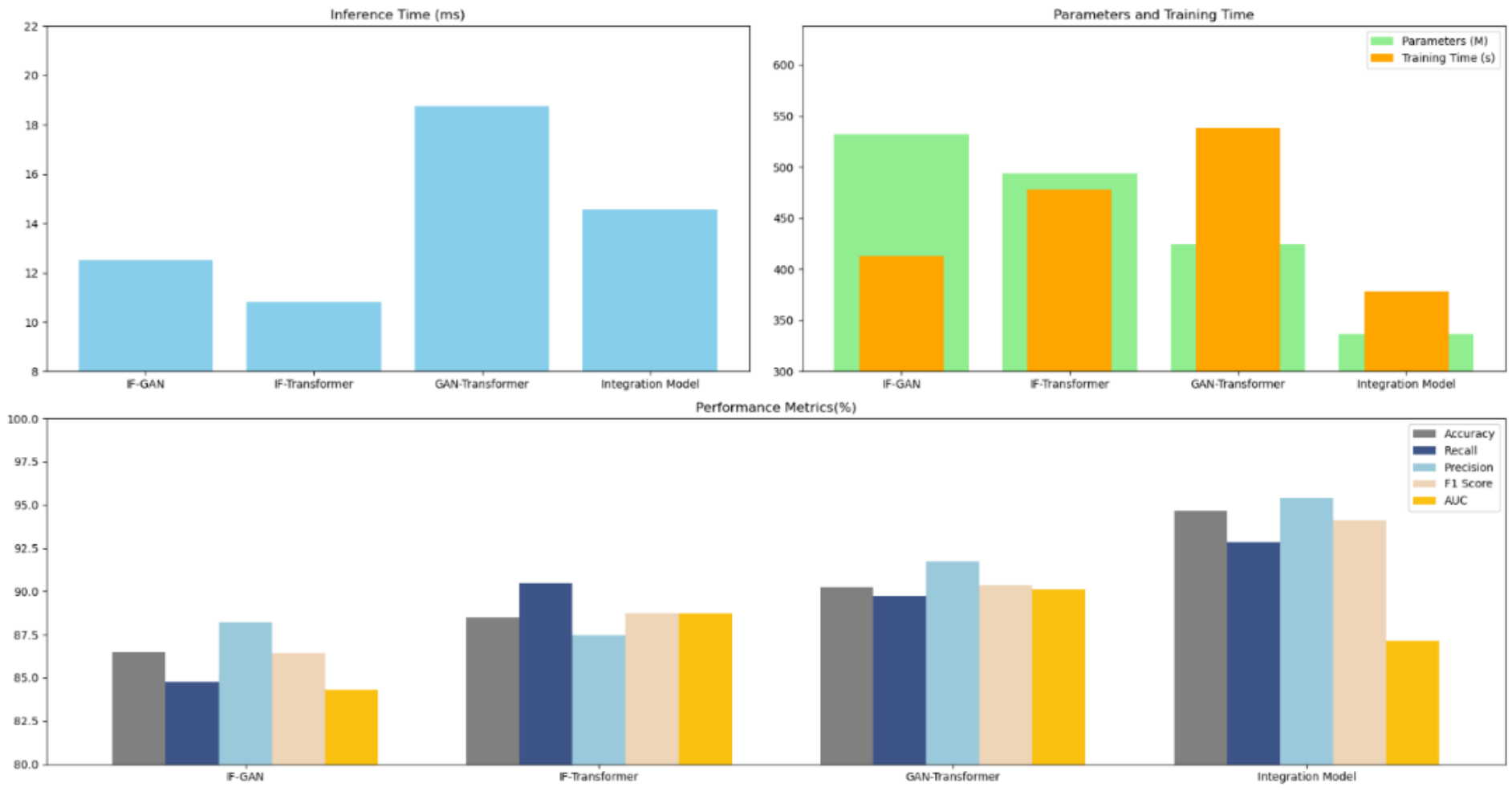

Figure 7. Visualization of Ablation Experiments.

## 5. Conclusion and Discussion

In this study, we propose a fusion model based on Isolation Forest, GAN and Transformer for anomaly detection and log analysis of time series data. Through extensive experimental evaluations and validation on multiple datasets with performance metrics, we demonstrate the effectiveness of the model. Our fusion model achieves significant improvements in the tasks of capturing anomalous behaviors in time-series data and log analysis compared to traditional approaches. In addition, we conducted model performance analysis to demonstrate its broad applicability across different datasets and tasks. The experimental phase is the key highlight of our research, where we deeply explore the performance of the model fusing Isolation Forest, GAN, and Transformer across multiple datasets and different application scenarios. Our experimental results show that the fused model significantly outperforms traditional approaches in anomaly detection and log analysis tasks. Specifically, we observe that on multiple datasets, the model not only improves the accuracy of anomaly detection, but also significantly reduces the false alarm rate, which is important for reducing false alarms. Meanwhile, the model performs well in log analysis, quickly identifying anomalous behaviors and helping enterprises detect potential problems in advance. These experimental results provide strong support for our model in real-world applications and demonstrate its potential in the real world.

Although our fusion model has made significant progress in anomaly detection and log analysis of time series data and demonstrated excellent performance, it still has some challenges and shortcomings that need to be further addressed. First, the performance of the model is limited by data quality and labeling. If the input data has noisy or incomplete features, the performance of the model may degrade. Second, the training and tuning process of the model requires larger computational resources and time, which may be challenging for some applications. And the interpretability of the models, although improved, still has room for improvement, especially for the explanation of complex abnormal behaviors and decision-making processes. Future work will focus on these aspects to further improve the comprehensiveness and usefulness of the model.

This research has made significant progress in the area of anomaly detection and log analysis of time series data, but this is only the beginning. Future work will focus on addressing the shortcomings of the model and further advancing the field of anomaly detection and log analysis of time series data. First, we plan to further investigate semi-supervised and unsupervised learning methods to reduce the need for labeled data and thus expand the applicability of the model. Second, we will continue to optimize the computational efficiency of the model to improve its efficiency and scalability so that it can handle larger data sizes and longer time series. Finally, we will explore new approaches to the interpretability of the model to help users better understand the decision-making process and results of the model, making the model's decisions more transparent and credible. In conclusion, the research in this paper not only provides new ideas and methods for the current analysis of time series data, but also provides a broad space for future research and application. We believe that this research will provide strong support for improving model performance, promoting technology development and solving practical problems in various fields.

## Author Contributions

Conceptualization, methodology, software, writing---original draft preparation, visualization, supervision, Shuzhan **Wang**; Validation, formal analysis, investigation, Ruxue **Jiang**; Methodology, Validation, writing---review and editing, Zhaoqi **Wang**; Data curation, writing---review and editing, Yan **Zhou**.

## Availability of Data and Materials

The data and materials used in this study are not currently available for public access. Interested parties may request access to the data by contacting the corresponding author.

## Conflicts of Interest

The author declares that the research was conducted in the absence of any commercial or financial relationships that could be construed as a potential conflict of interest.